\documentclass[10pt,twocolumn,letterpaper]{article}

\usepackage{wacv}      %

\usepackage[accsupp]{axessibility} %
\usepackage{graphicx}
\usepackage{amsmath}
\usepackage{amssymb}
\usepackage{booktabs}

\usepackage[pagebackref,breaklinks,colorlinks]{hyperref}

\usepackage[capitalize]{cleveref}
\crefname{section}{Sec.}{Secs.}
\Crefname{section}{Section}{Sections}
\Crefname{table}{Table}{Tables}
\crefname{table}{Tab.}{Tabs.}

\usepackage{booktabs}
\usepackage{microtype}
\usepackage{todonotes}
\usepackage{enumitem}
\usepackage{graphicx}
\usepackage{xspace}
\usepackage{xcolor}
\usepackage{amsmath}
\usepackage{multirow}
\usepackage{multicol}
\usepackage{caption}
\usepackage{hyperref}
\usepackage{subcaption}
\usepackage{colortbl}
\usepackage{framed}
\usepackage{comment}
\usepackage{footmisc}
\usepackage{listings}
\usepackage[normalem]{ulem}

\newcommand{\ourmodel}{Ladon\xspace}

\newcommand\YAMLcolonstyle{\color{red}\mdseries}
\newcommand\YAMLkeystyle{\color{black}\bfseries}
\newcommand\YAMLvaluestyle{\color{teal}\mdseries}
\lstdefinelanguage{yaml}{
    frame=single,
    keywords={true,false,null,y,n},
    keywordstyle=\color{darkgray}\bfseries,
    basicstyle=\YAMLkeystyle\small,
    sensitive=false,
    comment=[l]{!},
    columns=fullflexible,
    morecomment=[s]{/*}{*/},
    emph={!import_call},
    emphstyle=\color{brown},
    commentstyle=\color{brown}\ttfamily\bfseries,
    numberstyle=\color{blue},
    stringstyle=\YAMLvaluestyle,
    moredelim=[l][\color{orange}]{\&},
    moredelim=[l][\color{teal}\mdseries]{*},
    moredelim=**[il][\YAMLcolonstyle{:}\YAMLvaluestyle]{:},
    morestring=[b]',
    morestring=[b]"
}
\definecolor{mygreen}{rgb}{0,0.6,0}
\definecolor{mygray}{rgb}{0.5,0.5,0.5}
\definecolor{mymauve}{rgb}{0.58,0,0.82}
\lstdefinelanguage{mypython}{
    language=Python,
    frame=single,
    basicstyle=\ttfamily\scriptsize,
    comment=[l]{\#},
    keywordstyle=\ttfamily \color{blue},
    stringstyle=\color{teal}
}

\def\z{\mathbf{z}}
\def\zhat{\mathbf{\hat{z}}}
\def\h{\mathbf{h}}

\def\yhat{\mathbf{\hat{y}}}
\def\x{\mathbf{x}}
\def\xhat{\mathbf{\hat{x}}}

\def\encparams{\theta}
\def\encoder{f_\encparams}
\def\taskparams{\phi}
\def\taskmodule{g_\taskparams}

\def\wacvPaperID{461} %
\def\confName{WACV}
\def\confYear{2025}

\begin{document}

\title{A Multi-task Supervised Compression Model for Split Computing}

\author{Yoshitomo Matsubara \thanks{This work was done prior to joining Spiffy AI.}\\
University of California, Irvine\\
{\tt\small yoshitom@uci.edu}
\and
Matteo Mendula\\
University of Bologna\\
{\tt\small matteo.mendula@unibo.it}
\and
Marco Levorato\\
University of California, Irvine\\
{\tt\small levorato@uci.edu}
}

\maketitle

\begin{abstract}
Split computing ($\neq$ split learning) is a promising approach to deep learning models for resource-constrained edge computing systems, where weak sensor (mobile) devices are wirelessly connected to stronger edge servers through channels with limited communication capacity. State-of-the-art work on split computing presents methods for single tasks such as image classification, object detection, or semantic segmentation. The application of existing methods to multi-task problems degrades model accuracy and/or significantly increase runtime latency. In this study, we propose \ourmodel, the first multi-task-head supervised compression model for multi-task split computing.\footnote{Code and models are available at \url{https://github.com/yoshitomo-matsubara/ladon-multi-task-sc2} \label{fn:code}} Experimental results show that the multi-task supervised compression model either outperformed or rivaled strong lightweight baseline models in terms of predictive performance for ILSVRC 2012, COCO 2017, and PASCAL VOC 2012 datasets while learning compressed representations at its early layers. Furthermore, our models reduced end-to-end latency (by up to $95.4$\%) and energy consumption of mobile devices (by up to $88.2$\%) in multi-task split computing scenarios.
\end{abstract}

\section{Introduction}
\label{sec:introduction}
Learning compressed representations in a supervised manner has been empirically shown effective, specifically for training deep learning models for split computing (SC)~\cite{matsubara2022split}.
Split computing is an efficient distributed inference technique for resource-constrained edge computing systems, where either executing entire models on mobile devices or fully offloading tasks to cloud/edge servers may not be a feasible or optimal solution~\cite{matsubara2020head,matsubara2020neural}.
In split computing, the first layers of a deep learning model (\emph{i.e.}, encoder) are executed by the mobile device, while the remaining layers' computation are offloaded to an edge/cloud server via a wireless communication channel with a constrained capacity.
We remark that, different from distributed training such as federated learning~\cite{mcmahan2017communication} or split learning~\cite{vepakomma2018split}, models for split computing are trained offline and only executed over multiple machines at runtime.

A recent study~\cite{matsubara2023sc2} conducts comprehensive benchmark experiments of supervised compression for split computing (SC2) and empirically demonstrates that supervised compression outperforms input compression (\emph{e.g.}, codec-based and neural-based image compression methods for input images) and feature compression (\emph{e.g.}, introducing autoencoders~\cite{kingma2013auto} to intermediate layers) for single downstream tasks.
However, the study discusses single task scenarios only, and lacks latency and energy consumption evaluations from recent real-world embedded computers such as autonomous drones and smartphones for time-sensitive applications~\cite{di2019design}.

In this work, we propose \ourmodel, the first end-to-end multi-task supervised compression model for split computing that can serve multiple tasks in a fast, energy efficient manner.
Following the SC2 benchmark~\cite{matsubara2023sc2}, we assess the proposed model in terms of encoder size, compressed data size, and model accuracy, compared to popular lightweight models.
Besides the metrics from the SC2 benchmark, we consider end-to-end latency and energy consumption of mobile devices for multi-task split computing, where we execute multiple tasks given an input sample \emph{e.g.}, run image classification, object detection, and semantic segmentation tasks for an input image in resource-constrained edge computing systems.

Our contribution in this work is three-fold:
\begin{enumerate}
    \item We propose the first supervised compression model designed for multi-task split computing and empirically assess its predictive performance specifically for image classification (ILSVRC 2012), object detection (COCO 2017), and semantic segmentation (PASCAL VOC 2012) tasks~\cite{russakovsky2015imagenet,lin2014microsoft,everingham2012pascal}, using popular lightweight models as competitive baselines.
    \item Based on resource-constrained edge computing systems, we evaluate average end-to-end latency of the baselines and the \ourmodel model in multi-task scenarios. Our \ourmodel model takes advantage of its single, lightweight multi-task encoder and saves up to $95.4$\% of the end-to-end latency for the baselines. %
    \item Using NVIDIA Jetson devices, we assess energy consumption of the baseline and proposed methods on the mobile devices in the same multi-task scenarios. Our approach successfully reduces the energy consumption by $65.0 - 88.2$\%.
\end{enumerate}

\section{Preliminaries}
\label{sec:preliminary}
In split computing~\cite{matsubara2022split}, resource-constrained edge computing systems consist of three components: 1) a weak mobile (local) device; 2) a wireless communication link with limited capacity between 1) and 3); and 3) a compute-capable edge/cloud server.
Local devices have limited computing resources and battery constraints \emph{e.g.}, NVIDIA Jetson devices.
The wireless communication link can achieve a limited (and volatile) data rate due to power and bandwidth contraints and impaired propagation.
To make an example, a low-power communication technology, LoRa, has a maximum data rate of 37.5 Kbps.

In this setting, from an execution perspective it is preferable for the mobile devices to offload as much computing load as possible to the edge server, thus minimizing local computing (LC) load and energy consumption.
However, in order to achieve low end-to-end latency, it is also essential to reduce the size of the data transmitted over the limited wireless channel to support offloading.
Intuitively, the more we compress data, the more we sacrifice a model's predictive performance \emph{e.g.}, model accuracy.
In general, it is easier to aggressively compress the data at late layers of a deep learning model (\emph{e.g.}, the penultimate layer) without significant loss of model accuracy, which, however, results in high encoding cost at the mobile devices.
This results in a known three-way tradeoff between encoding cost, data size, and model accuracy in the context of supervised compression for split computing (SC2)~\cite{matsubara2023sc2}.

\section{Related Work}
\label{sec:related_work}
Split computing approaches have been actively studied in different research communities~\cite{matsubara2022split}, and compared with local computing and full offloading baselines for resource-constrained edge computing systems.
The former runs the full model on a weaker local device, and the latter transmits sensor data (\emph{e.g.}, an image) to a stronger edge server via limited wireless connection channels and runs the model.
Note that split computing is different from split learning~\cite{vepakomma2018split}, as split computing only involves online inference of offline trained models.

A multi-task model for collaborative inference, called Chimera, is proposed in~\cite{nimi2022chimera}, where two variants of the model are discussed.
One of the models serves two different image classification tasks (gender and attribute) for $32\times32$ pixel face images from the CelebA dataset~\cite{liu2015deep}, and the other serves two different pixel-wise predictions (semantic segmentation and depth estimation) for NYUv2 dataset~\cite{silberman2012indoor}.
However, this approach is designed under the implicit assumption that a reference model to be trained ``naturally'' contains at least one layer whose output data size is smaller than the input data size, which is called a bottleneck.
Such a strong assumption not only limits data compression gain, but also implies non-trivial computing cost on weak mobile devices as effective splittable layers only occur later in computer vision models.
In fact, it is shown in their study that their approach requires to execute at least the first 34 layers in their model to produce smaller data than the input image. 

The entropic student~\cite{matsubara2022supervised} is a supervised compression method that combines concepts of neural image compression and knowledge distillation for split computing.
The method offers a sharable learned encoder and can serve multiple computer vision tasks.
A comprehensive benchmark study on supervised compression for split computing (SC2)~\cite{matsubara2023sc2} empirically shows that the entropic student is currently the best SC2 approach for three challenging computer vision tasks: ILSVRC 2012 (image classification)~\cite{russakovsky2015imagenet}, COCO 2017 (object detection)~\cite{lin2014microsoft}, and PASCAL VOC 2012 (semantic segmentation)~\cite{everingham2012pascal}.
We consider the approach as the strongest baseline to compare with this study from the perspective of the three-way tradeoff illustrated earlier.
While its learned encoder can serve multiple tasks, the subsequent modules are independently trained on each of the target tasks and is not optimized as a multi-task model at runtime.
This results in a large computing load at the edge server if the input image is to be analyzed to extract different outputs.

\section{\ourmodel ~- Proposed Approach -}
\label{sec:proposed_method}

\begin{figure}[t]
    \centering
    \includegraphics[width=0.99\linewidth]{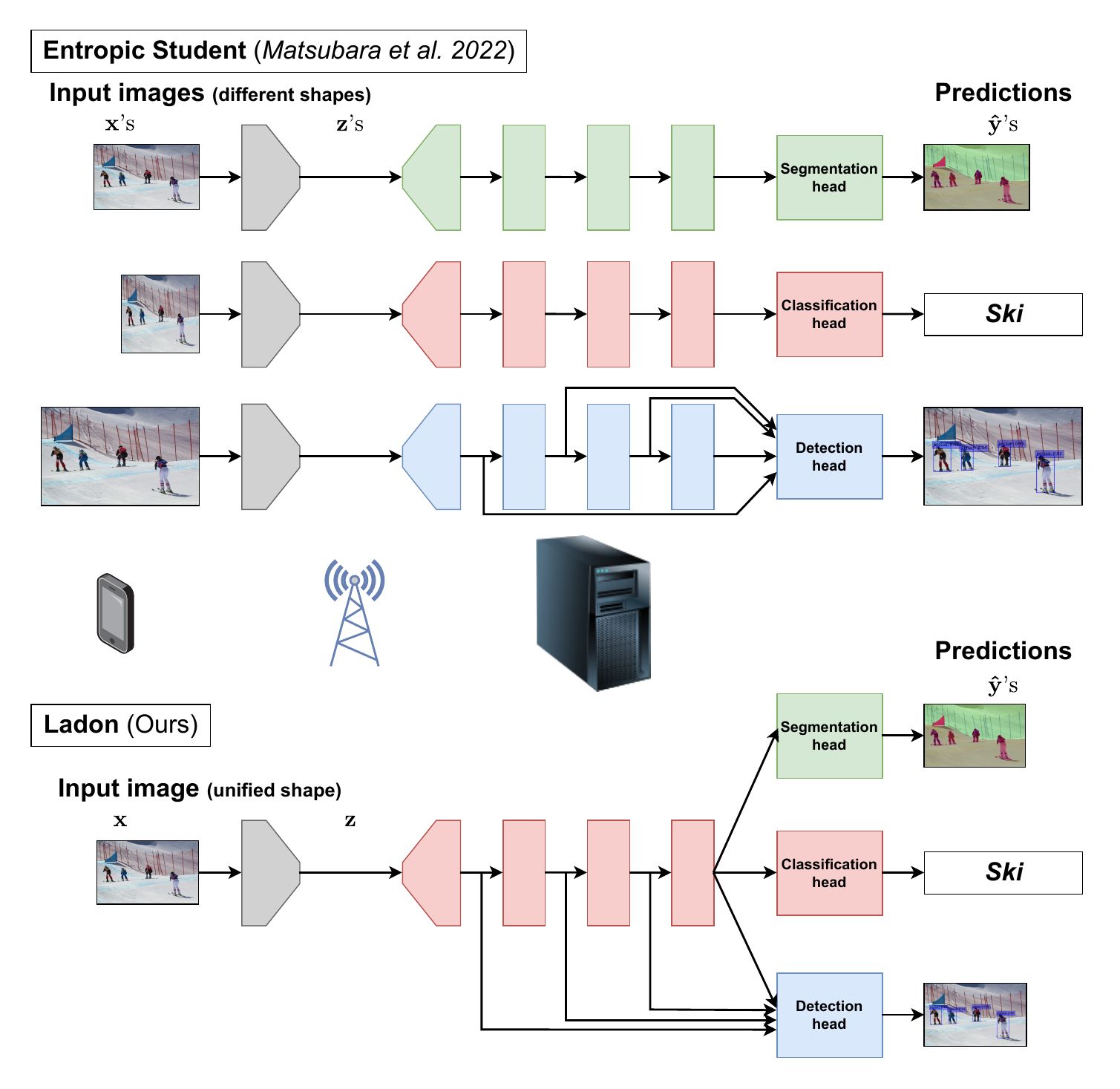}
    \vspace{-0.5em}
    \caption{Entropic student (top) vs. our \ourmodel (bottom) in multi-task scenario. Gray module (encoder) is trained in a task-agnostic way. \textcolor{red}{Red}, \textcolor{teal}{green}, and \textcolor{blue}{blue} modules are trained for \textcolor{red}{image classification}, \textcolor{teal}{semantic segmentation}, and \textcolor{blue}{object detection} tasks respectively. Entropic student's encoder serves multiple downstream tasks, but all its modules except the encoder are trained independently, using different image preprocessing pipelines such as resizing and cropping. Our \ourmodel model shares a single image preprocessing pipeline and most of its parameters (except those of task-specific heads) across the downstream tasks at run time. While Entropic Student is designed to run three separate inferences per image, the \ourmodel model runs a single inference to serve the three tasks.}
    \label{fig:es_vs_ladon}
\end{figure}

In this section, we describe our proposed \ourmodel model, highlighting the differences from the entropic student~\cite{matsubara2022supervised} in Fig.~\ref{fig:es_vs_ladon}.
Both the entropic student and our \ourmodel models are supervised compression models for split computing.
Supervised compression is defined as learning compressed representations for supervised downstream tasks~\cite{matsubara2023sc2}.

\subsection{Problem Formulation}
\label{subsec:problem_formulation}
In general, a supervised compression model is trained for a single task (\emph{e.g.}, image classification)~\cite{eshratifar2019bottlenet,matsubara2019distilled,singh2020end,matsubara2022supervised} and has a deterministic mapping $\x \mapsto \zhat \mapsto \yhat$, where $\x$, $\zhat$, and $\yhat$ indicate the input data, compressed representation, and prediction, for a given supervised downstream task.
Note that $\zhat$ contains more relevant information with supervised tasks rather than information to reconstruct the original input data $\x$, which is different from neural image compression~\cite{balle2017end,balle2018variational} that learns a compressed representation of the input data $\x$ and produces the reconstructed data $\xhat$ such that $\x \approx \xhat$ in an unsupervised manner.

We define $\encoder$ as an encoder with learned parameters $\encparams$ to transform an input image $\x$ to the corresponding compressed data $\zhat$ \emph{i.e.}, $\encoder(\x) = \zhat$.
Similarly, we define $\taskmodule$ as the remaining layers in the supervised compression model to produce a task-specific prediction $\yhat$ including task-specific learned parameters $\taskparams$ and the encoded data $\z$ from the local device \emph{i.e.}, $\taskmodule(\zhat) = \yhat$.

At runtime, the encoder $\encoder$ runs on the mobile device and then transmits encoded data $\zhat$ to an edge server via a wireless channel.
Once the edge server receives the encoded data $\zhat$, the remaining layers (including the decoder) $\taskmodule$ are executed on the edge server to produce the prediction $\yhat$.

\paragraph{Shared encoder vs. end-to-end multi-task model}
As shown in Fig.~\ref{fig:es_vs_ladon} (top), the Entropic Student method trains a single encoder that serves multiple supervised tasks, and then trains the subsequent modules for each of the individual tasks.
Those models may share some of the network architectures between the tasks (\emph{e.g.}, ResNet-50~\cite{he2016deep} as a backbone in~\cite{matsubara2022supervised}), but their parameters are separately optimized for each of the tasks.
Thus, in multi-task scenarios, the entropic student method requires multiple models to be deployed on the edge server, which increases memory footprint and computing cost.
Our \ourmodel model addresses the limitation and shares not only the network architecture, but also learned parameters of a backbone model between the tasks.
As illustrated in Fig.~\ref{fig:es_vs_ladon} (bottom), the \ourmodel model is designed to minimize task-specific components and enable to simultaneously serve multiple tasks.
\emph{I.e.}, our model produces multi-task predictions given an input image at runtime.

\paragraph{Unified preprocessing}
Notably, different computer vision tasks have different requirements during the image preprocessing phase (see Table~\ref{table:baseline_models} and Fig.~\ref{fig:es_vs_ladon}).
For image classification tasks, a popular preprocessing at runtime is a combination of resizing an image to fixed patch size and center-cropping the resized image with a smaller fixed patch size (\emph{e.g.}, $224\times224$ pixels).
Object detection and semantic segmentation tasks usually prefer more flexible resizing (keeping height and width ratio of the image) and do not use cropping since model predictions such as bounding box and pixel-wise predictions need to be overlaid on the input image.
As illustrated in Fig.~\ref{fig:es_vs_ladon}, the entropic student method trains different subsequent modules with a unique preprocessing pipeline for each of the tasks while our \ourmodel model is optimized with a unified preprocessing pipeline so that the model can simultaneously serve multiple tasks given an input image without model accuracy degradation.

\subsection{Model Implementations}
\label{subsec:implementations}
We implement our \ourmodel model using two backbone models pretrained on the ILSVRC 2012 dataset: ResNet-50 (\#params: 25.6M)~\cite{he2016deep} and ResNeSt-269e (\#params: 110M)~\cite{zhang2022resnest}.
Specifically, we modify each of the backbone models and replace its first layers (including its first residual block) with supervised compression encoder-decoder architectures~\cite{matsubara2022supervised}.
The modified backbone architectures serve an image classification task for ILSVRC 2012 (ImageNet) dataset and are reused as part of Faster R-CNN with feature pyramid network (FPN)~\cite{ren2015faster,lin2017feature} and DeepLabv3~\cite{chen2017rethinking} for object detection and semantic segmentation tasks (COCO 2017 and PASCAL VOC 2012 datasets).
Our code repository and trained PyTorch~\cite{paszke2019pytorch} models are publicly available.\textsuperscript{\ref{fn:code}}

\subsection{Training}
\label{subsec:training}

\begin{table*}
    \centering
    \small
    \begin{tabular}{l|llr}
        \toprule
        \multicolumn{1}{c|}{\bf Device} & \multicolumn{1}{c}{\bf CPU}  & \multicolumn{1}{c}{\bf GPU} & \multicolumn{1}{c}{\bf RAM} \\
        \hline\midrule
        NVIDIA Jetson Nano & Quad-core ARM Cortex-A57 & 128-core NVIDIA Maxwell & 4 GB \\
        NVIDIA Jetson Xavier NX & 6-core NVIDIA Carmel 64-bit ARMv8.2 & 384-core NVIDIA Volta & 8 GB \\
        Laptop computer & Intel Core i9-11950H & NVIDIA RTX A500 Mobile & 32 GB \\
        \bottomrule
    \end{tabular}
    \caption{Device specifications}
    \label{table:devices}
\end{table*}

\begin{table*}
    \centering
    \small
    \bgroup
    \setlength{\tabcolsep}{0.6em}
    \begin{tabular}{crlrrlr}
        \toprule
        {\bf Task} & \multicolumn{1}{c}{\bf Index}  & \multicolumn{1}{c}{\bf Name} & \multicolumn{1}{c}{\bf \#Params.}  & \multicolumn{1}{c}{\bf Size}  & \multicolumn{1}{c}{\bf Input Shape}& \multicolumn{1}{c}{\bf Task metric} \\
        \hline\midrule
        \multirow{4}{*}{ILSVRC 2012 (IC)} & 1 & MobileNetV3 (MNv3) Large 1.0~\cite{howard2019searching} & 5.5M & 21.0 MB & (3, 224, 224) & Acc. 74.0 \\
        & 2 & MobileNetV2~\cite{sandler2018mobilenetv2} & 3.5M & 13.5 MB & (3, 224, 224) & Acc. 71.9 \\
        & 3 & MNASNet 1.3~\cite{tan2019mnasnet} & 6.3M & 24.2 MB & (3, 224, 224) & Acc. 76.5 \\
        & 4 & MNASNet 0.5~\cite{tan2019mnasnet} & 2.2M & 8.5 MB & (3, 224, 224) & Acc. 67.7 \\
        \midrule
        \multirow{3}{*}{COCO 2017 (OD)} & 5 & SSD300 w/ VGG-16~\cite{liu2016ssd} & 35.6M & 136 MB & (3, 300, 300) & mAP 25.1 \\
        & 6 & SSD Lite w/ MNv3~\cite{howard2019searching} & 5.2M & 20.0 MB & (3, 320, 320) & mAP 21.3 \\
        & 7 & Faster R-CNN w/ MNv3 + FPN~\cite{ren2015faster} & 19.4M & 74.1 MB & (3, 320+, 320+) & mAP 22.8 \\
        \midrule
        \multirow{2}{*}{PASCAL VOC 2012 (SS)} & 8 & DeepLabv3 w/ MNv3~\cite{chen2017rethinking} & 11.0M & 42.2 MB & (3, 416+, 416+) & mIoU 73.4 \\
        & 9 & LRASSP w/ MNv3~\cite{howard2019searching} & 3.2M & 12.4 MB & (3, 416+, 416+) & mIoU 73.3 \\
        \bottomrule
    \end{tabular}
    \caption{Accuracy [\%], mean average precision (mAP) [\%], and mean intersection over union (mIoU) of baseline models for ILSVRC 2012 (IC: image classification), COCO 2017 (OD: object detection), and PASCAL VOC 2012 (SS: semantic segmentation), respectively}
    \label{table:baseline_models}
    \egroup
\end{table*}

We introduce three steps to train our \ourmodel model.
We describe its hyperparameters in the supplementary material.

\paragraph{Step 1: Pre-training encoder-decoder}
Following~\cite{matsubara2022supervised}, we begin by training the modified backbone model using the original pretrained backbone model as a teacher model for the ILSVRC 2012 dataset.
We minimize a loss function
\begin{eqnarray}
    \mathcal{L_\text{pre}}(\x) = \underbrace{\sum_{i \in \mathbb{I}}|| \h_i^\text{t}(\x) - \h_i^\text{s}(\x) ||_{2}^{2}}_\text{distortion} - \beta \underbrace{\log p_\phi(f_\theta(\x) + \epsilon)}_{\text{rate}}\\
    \epsilon \sim \text{Unif}(\textstyle{-\frac{1}{2},\frac{1}{2}}), \nonumber
    \label{eq:1st_loss}
\end{eqnarray}
\noindent where $\mathbb{I}$ is a set of indices for teacher-student layer pairs.
$f_\theta$ and $g_\phi$ indicate encoder and decoder of a student model, respectively.
$p_\psi$ is a prior probability model of quantized representations (the entropy model) and used for both the encoder and decoder~\cite{balle2018variational}.
$\h_i^{t}(\x)$ and $\h_i^{s}(\x)$ indicate embeddings of the $i$-th pair of teacher and student layers for a given model input $\x$.
$\beta$ is a hyperparameter that controls a rate-distortion tradeoff (\emph{e.g.}, compressed data size and predictive performance).
Note that for our \ourmodel model, $\h_1^{s}(\x)$ in training is an output of its decoder \emph{i.e.}, $\h_1^{s}(\x) = g_{\phi}\left(f_{\theta}\left(\mathbf{x}\right) + \epsilon \right)$.

\paragraph{Step 2: Fine-tuning decoder and subsequent modules}
Once the pre-training of the encoder-decoder module is completed, we fix the parameters of the encoder and fine-tune the decoder and the subsequent modules in the modified backbone models.
The purpose of this fine-tuning step is to make their parameters reusable for relevant downstream tasks.
In this study, we fine-tune the modules by a standard knowledge distillation~\cite{hinton2014distilling} on the ILSVRC 2012 dataset given that ImageNet pre-training is empirically demonstrated to speed up convergence of CNN models on the target task~\cite{he2019rethinking}.

\paragraph{Step 3: Fine-tuning other task-specific modules}
Following the step 2, we freeze all the parameters of the model and introduce other task-specific heads as the \ourmodel's additional branches (see Fig.~\ref{fig:es_vs_ladon}).
Specifically, we attach object detection and semantic segmentation modules to the model and fine-tune the newly attached task-specific heads.
Due to the nature of supervised compression (Section~\ref{subsec:problem_formulation}), the additional tasks should be related to the first supervised task.

\section{Experiments}
\label{sec:experiments}
We conduct comprehensive experiments using various computing devices and heterogeneous lightweight baselines.

\subsection{Experimental Settings}
\label{subsec:experimental_settings}

Following~\cite{matsubara2023sc2}, we consider the three challenging computer vision datasets for predictive performance evaluations: ILSVRC 2012\footnote{\url{https://image-net.org/challenges/LSVRC/2012/}} (image classification)~\cite{russakovsky2015imagenet}, COCO 2017\footnote{\url{https://cocodataset.org/\#detection-2017}} (object detection)~\cite{lin2014microsoft}, and PASCAL VOC 2012\footnote{\url{http://host.robots.ox.ac.uk/pascal/VOC/voc2012/}} (semantic segmentation)~\cite{everingham2012pascal}.
For end-to-end latency evaluations, we simulate resource-constrained edge computing systems, considering two devices with unbalanced computing resources (one weak mobile device and one strong edge server) and a limited wireless communication link (data rate: 100 Kbps) or LoRa~\cite{samie2016iot} - a protocol for low-power communications, and its maximum data rate is 37.5 Kbps.
Table~\ref{table:devices} summarizes computing devices we consider for end-to-end latency evaluations.
Making a few pairs of the computing devices, we assess end-to-end latency in multi-task scenarios for each of the baseline methods and our proposed approach.

\subsection{Baselines}
\label{subsec:baselines}

Table~\ref{table:baseline_models} summarizes lightweight single-task models that we compare with our \ourmodel (multi-task model) in terms of predictive performance for image classification, object detection, and semantic segmentation tasks.
To highlight differences from our end-to-end multi-task model at runtime (see Fig.~\ref{fig:es_vs_ladon}), we consider Entropic Student~\cite{matsubara2022supervised}, which uses a shared encoder and separate image preprocessing pipelines for different tasks, as an additional baseline when discussing end-to-end latency and local energy consumption.

\subsection{Evaluation Metrics}
\label{subsec:eval_metrics}

Our evaluations are based on the benchmark framework for split computing in~\cite{matsubara2023sc2}, which adopts three key metrics: 1) encoder size, 2) compressed data size, and 3) model accuracy.
Additionally, we consider end-to-end latency and energy consumption of mobile devices in multi-task scenarios.
To estimate energy consumption, we employ Energon~\cite{mendula2024furcifer}, a power monitoring tool compliant to the Prometheus standard~\cite{prometheus}.
Energon integrates with the built-in power monitoring registries of Jetson devices and x86 Intel-based laptop architectures.
To convert instantaneous power consumption into Joules, we use Simpson's rule for numerical integration~\cite{tallarida1987area,agiollo2024enea} over the same number of inferences for each experimental configuration.
We warm up the devices before conducting latency and power consumption experiments to stabilize GPU behavior~\cite{dao2014performance}.
Using the five different evaluation metrics, we comprehensively discuss the proposed method with respect to our baselines.

\section{Results}
\label{sec:results}

This section discusses results of the experiments.
Section~\ref{subsec:static_metrics} presents the evaluation results in terms of static SC2 metrics: task-specific rate-distortion tradeoff (compressed data size and predictive performance) and total encoder size for multi-task scenarios in split computing.
Sections~\ref{subsec:e2e_latency} and~\ref{subsec:local_energy_consumption} show the end-to-end latency and the energy consumption for local devices, respectively.

\begin{figure}[t]
    \centering
    \includegraphics[width=0.975\linewidth]{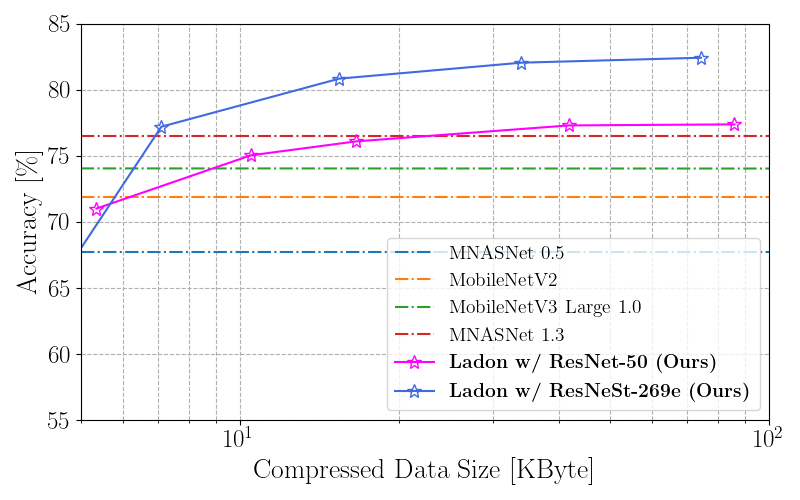}
    \vspace{-1em}
    \caption{ILSVRC 2012: tradeoff between compressed data size and model accuracy}
    \label{fig:rd_tradeoff_ic}
    \vspace{0.5em}
    \includegraphics[width=0.975\linewidth]{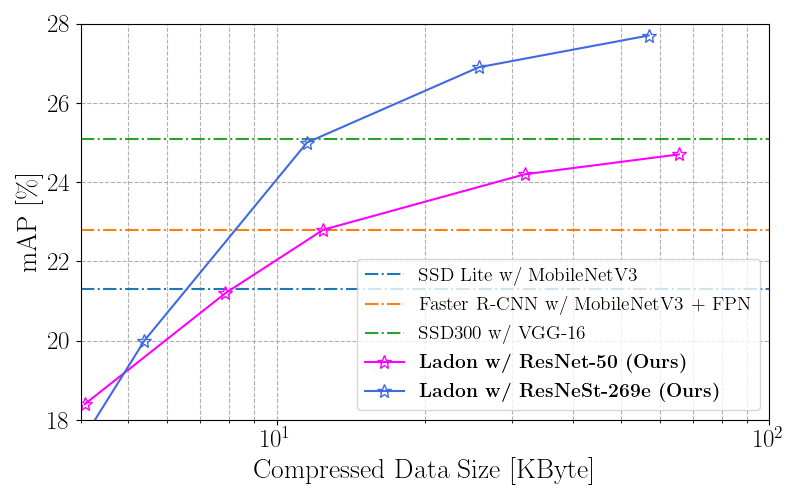}
    \vspace{-1em}
    \caption{COCO 2017: tradeoff between compressed data size and mean average precision (mAP)}
    \label{fig:rd_tradeoff_od}
    \vspace{0.5em}
    \includegraphics[width=0.975\linewidth]{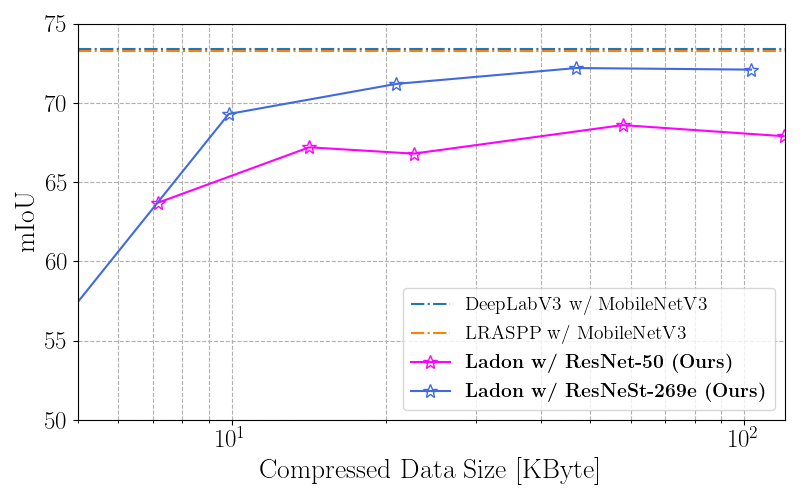}
    \vspace{-1em}
    \caption{PASCAL VOC 2012: tradeoff between compressed data size and mean intersection over union (mIoU)}
    \label{fig:rd_tradeoff_ss}
\end{figure}

\begin{table}[t]
    \centering
    \small
    \begin{tabular}{l|rrr}
        \toprule
        \multicolumn{1}{c|}{\bf Config.} & \multicolumn{1}{c}{\bf IC model}  & \multicolumn{1}{c}{\bf OD model}& \multicolumn{1}{c}{\bf SS model} \\
        \hline\midrule
        LC 1 & 1 & 5 & 8 \\
        LC 2 & 2 & 5 & 8 \\
        LC 3 & 3 & 5 & 8 \\
        LC 4 & 4 & 5 & 8 \\
        LC 5 & 1 & 6 & 8 \\
        LC 6 & 2 & 6 & 8 \\
        LC 7 & 3 & 6 & 8 \\
        LC 8 & 4 & 6 & 8 \\
        LC 9 & 1 & 7 & 8 \\
        LC 10 & 2 & 7 & 8 \\
        LC 11 & 3 & 7 & 8 \\
        LC 12 & 4 & 7 & 8 \\
        LC 13 & 1 & 5 & 9 \\
        LC 14 & 2 & 5 & 9 \\
        LC 15 & 3 & 5 & 9 \\
        LC 16 & 4 & 5 & 9 \\
        LC 17 & 1 & 6 & 9 \\
        LC 18 & 2 & 6 & 9 \\
        LC 19 & 3 & 6 & 9 \\
        LC 20 & 4 & 6 & 9 \\
        LC 21 & 1 & 7 & 9 \\
        LC 22 & 2 & 7 & 9 \\
        LC 23 & 3 & 7 & 9 \\
        LC 24 & 4 & 7 & 9 \\
        \midrule
        SC 1 & ES-IC 1 & ES-OD 1 & ES-SS 1 \\
        SC 2 & ES-IC 2 & ES-OD 2 & ES-SS 2 \\
        SC 3 & ES-IC 3 & ES-OD 3 & ES-SS 3 \\
        SC 4 & ES-IC 4 & ES-OD 4 & ES-SS 4 \\
        SC 5 & ES-IC 5 & ES-OD 5 & ES-SS 5 \\
        SC 6 & ES-IC 6 & ES-OD 6 & ES-SS 6 \\
        \midrule
        Ours 1 & \multicolumn{3}{l}{\ourmodel (ResNet-50) w/ $\beta = 0.32$} \\
        Ours 2 & \multicolumn{3}{l}{\ourmodel (ResNet-50) w/ $\beta = 1.28$} \\
        Ours 3 & \multicolumn{3}{l}{\ourmodel (ResNet-50) w/ $\beta = 5.12$} \\
        Ours 4 & \multicolumn{3}{l}{\ourmodel (ResNet-50) w/ $\beta = 10.24$} \\
        Ours 5 & \multicolumn{3}{l}{\ourmodel (ResNet-50) w/ $\beta = 20.48$} \\
        Ours 6 & \multicolumn{3}{l}{\ourmodel (ResNeSt-269e) w/ $\beta = 0.32$} \\
        Ours 7 & \multicolumn{3}{l}{\ourmodel (ResNeSt-269e) w/ $\beta = 1.28$} \\
        Ours 8 & \multicolumn{3}{l}{\ourmodel (ResNeSt-269e) w/ $\beta = 5.12$} \\
        Ours 9 & \multicolumn{3}{l}{\ourmodel (ResNeSt-269e) w/ $\beta = 10.24$} \\
        Ours 10 & \multicolumn{3}{l}{\ourmodel (ResNeSt-269e) w/ $\beta = 20.48$} \\
        \bottomrule
    \end{tabular}
    \caption{End-to-end multi-task latency evaluation configurations. LC model indices in Table~\ref{table:baseline_models}. ES-IC/OD/SS: Entropic Student (ResNet-50 / Faster R-CNN w/ ResNet-50 and FPN / DeepLabv3 w/ ResNet-50). ES-\text{*} $\{1, \cdots, 5\}$ are Entropic Student models trained with $\beta = 0.32, 0.64, 1.28, 2.56,$ and $5.12$, respectively~\cite{matsubara2023sc2}.}
    \label{table:e2e_configs}
\end{table}

\subsection{Static SC2 Metrics}
\label{subsec:static_metrics}

Figures~\ref{fig:rd_tradeoff_ic} -~\ref{fig:rd_tradeoff_ss} show task-specific rate-distortion tradeoffs for our \ourmodel models referring to lightweight models as local computing baselines.
We confirm that \ourmodel models outperform many of the models designed for mobile execution in terms of task-specific prediction performance.
For the semantic segmentation task, the \ourmodel models achieve comparable predictive performance with respect to the baselines.
It is understandable that the \ourmodel models encounter a more difficult challenge in the semantic segmentation task due to the lossy compression of the representation as the task requires pixel-level predictions while image classification and object detection are image-level and box-level predictions.

Besides the rate-distortion tradeoff, the SC2 benchmark~\cite{matsubara2023sc2} suggests encoder size as an additional static evaluation metric.
It is notable that the encoder sizes of our \ourmodel models are only $0.543 - 0.935$ MB, which are approximately $0.268 - 2.29$\% of total model size for multi-task local computing configurations (Tables~\ref{table:baseline_models} and~\ref{table:e2e_configs}).

\subsection{End-to-end Latency}
\label{subsec:e2e_latency}

With 30 different baseline configurations and 10 configurations for \ourmodel models (see Table~\ref{table:e2e_configs}), we discuss end-to-end latency evaluations based on devices listed in Table~\ref{table:devices}.

\begin{figure*}[t]
    \centering
    \includegraphics[width=0.95\linewidth]{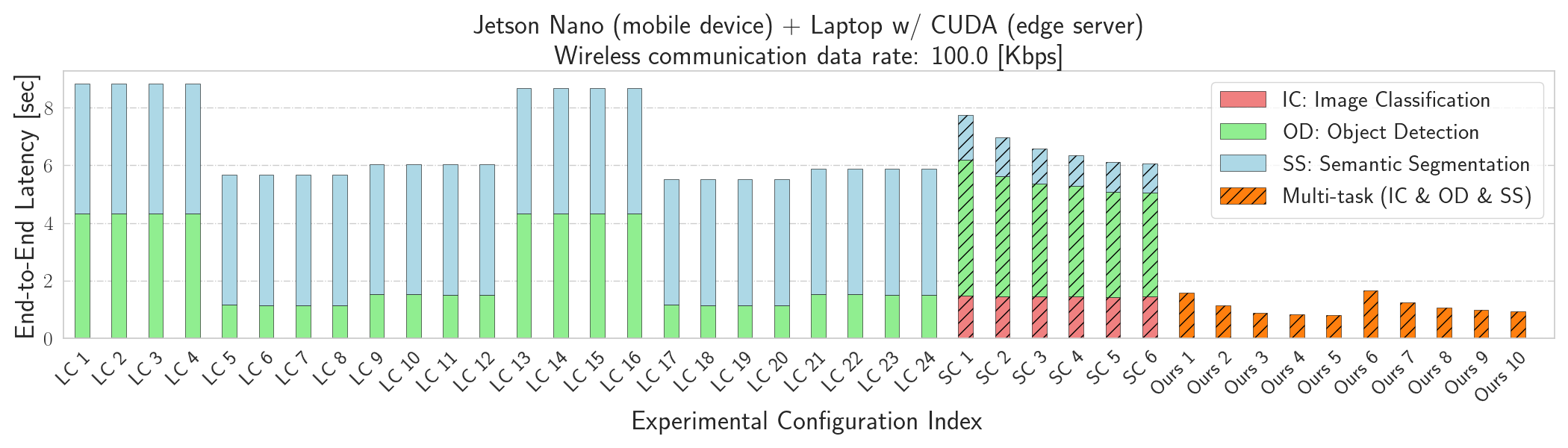}
    \label{fig:jetson_nano-laptop_w_cuda-100.0kbps}
    \vspace{1em}
    \includegraphics[width=0.95\linewidth]{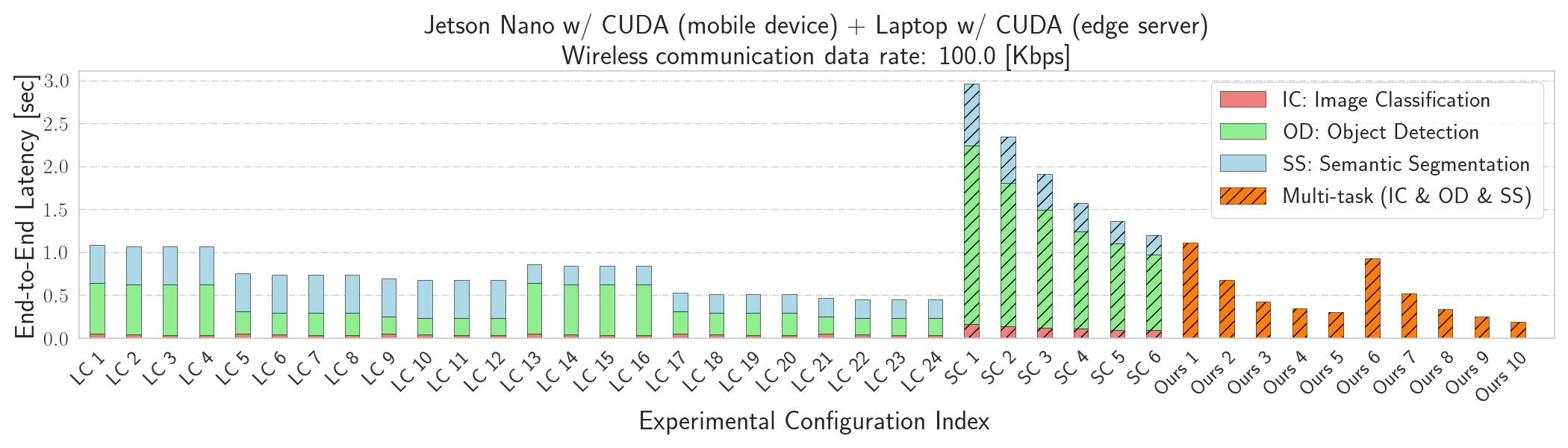}
    \vspace{-1em}
    \caption{End-to-end latency for Jetson Nano (mobile device) and laptop with CUDA (edge server). Top/bottom: local computing without/with CUDA.}
    \label{fig:jetson_nano_w_cuda-laptop_w_cuda-100.0kbps}
    \vspace{1em}
    \includegraphics[width=0.95\linewidth]{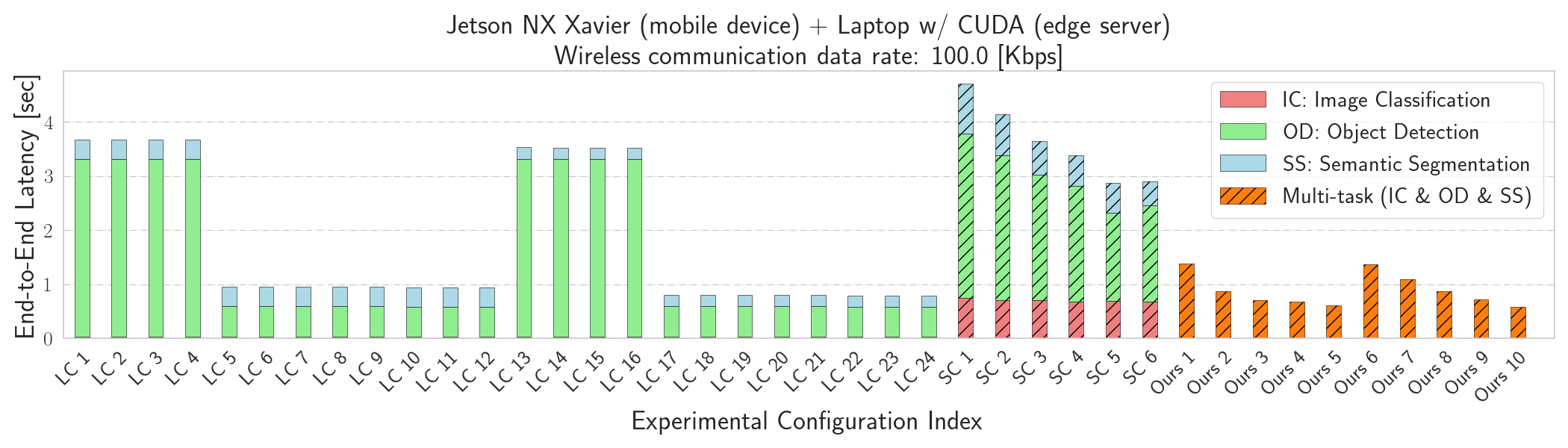}
    \label{fig:jetson_nx-laptop_w_cuda-100.0kbps}
    \vspace{1em}
    \includegraphics[width=0.95\linewidth]{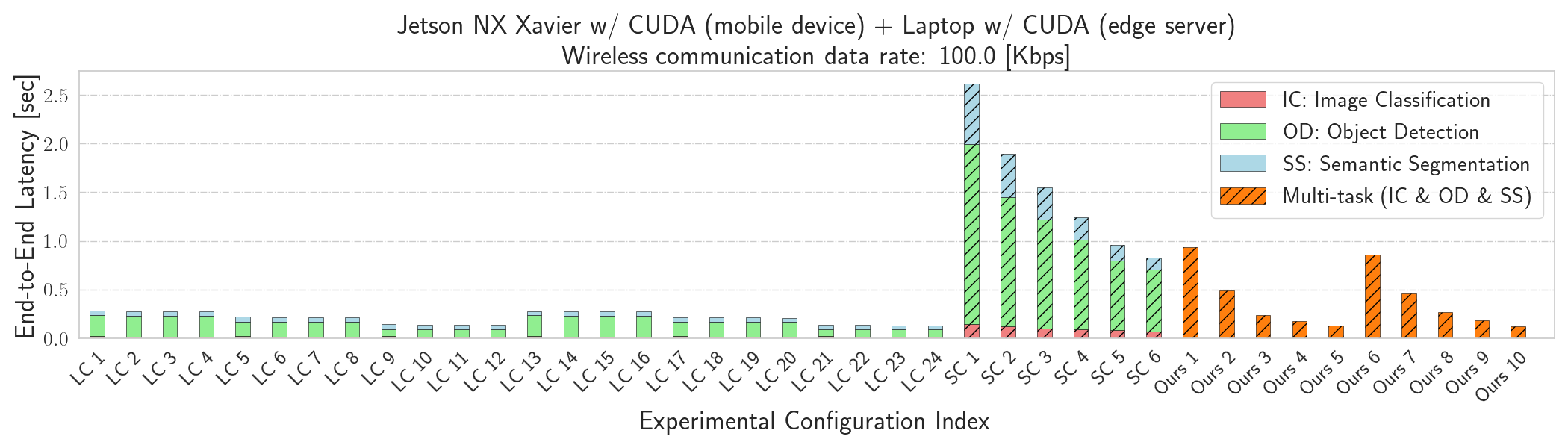}
    \vspace{-1em}
    \caption{End-to-end latency for Jetson NX Xavier (mobile device) and laptop with CUDA (edge server). Top/bottom: local computing without/with CUDA.}
    \label{fig:jetson_nx_w_cuda-laptop_w_cuda-100.0kbps}
\end{figure*}

Figure~\ref{fig:jetson_nano_w_cuda-laptop_w_cuda-100.0kbps} show the end-to-end latency when using Jetson Nano (top: CUDA OFF, bottom: CUDA ON) and laptop as mobile device and edge server, respectively.
For local computing (LC) baselines, their image classification delays are relatively small with respect to those of object detection and semantic segmentation models considered in this study.
When CUDA was turned off, split computing (SC) baselines slightly outperformed local computing baselines (up to $31.3$\% reduction), and our multi-task models reduced the end-to-end latency by up to $82.6$\% and $93.6$\% with respect to the LC and SC baselines respectively.
The local computing baselines took more advantage of CUDA than the split computing baselines did, but our models still achieved comparable or slightly improved end-to-end latency.
We confirmed similar trends in Fig.~\ref{fig:jetson_nx_w_cuda-laptop_w_cuda-100.0kbps} where we used Jetson NX Xavier, a stronger mobile device than Jetson Nano, which made the local computing baselines even stronger.
Yet, our approach reduced the end-to-end latency of the LC and SC baselines by up to $84.2$\% and $95.3$\%, respectively.

We refer readers to the supplementary material for additional evaluations that use LoRa data rate ($37.5$ Kbps).

\subsection{Local Energy Consumption}
\label{subsec:local_energy_consumption}

\begin{figure*}[t]
    \centering
    \includegraphics[width=0.975\linewidth]{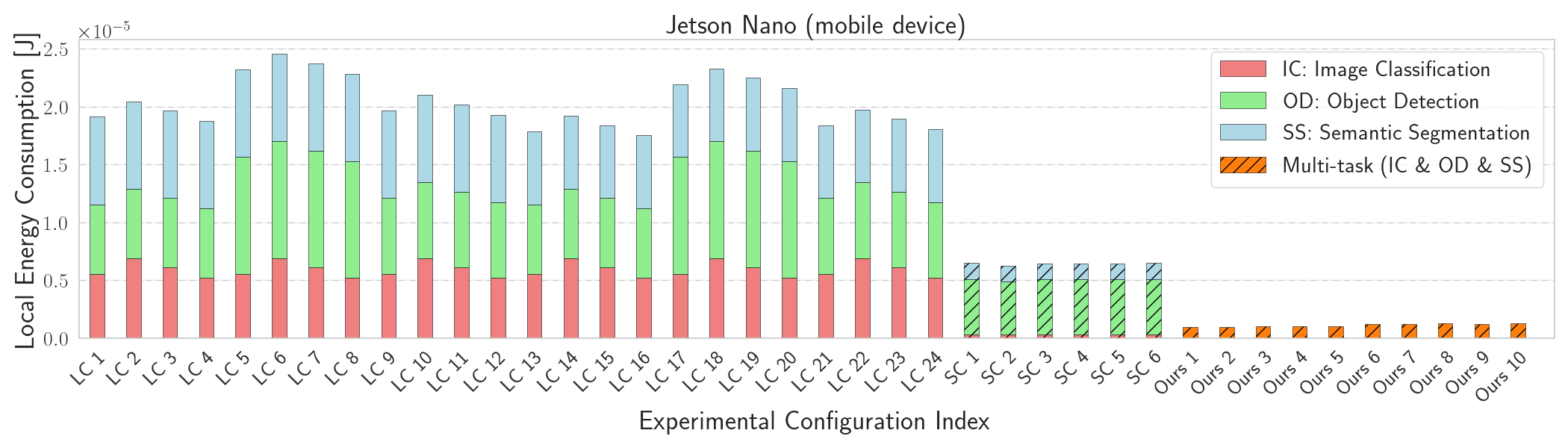}
    \label{fig:jetson_nano-energy_consumption}
    \vspace{1em}
    \includegraphics[width=0.975\linewidth]{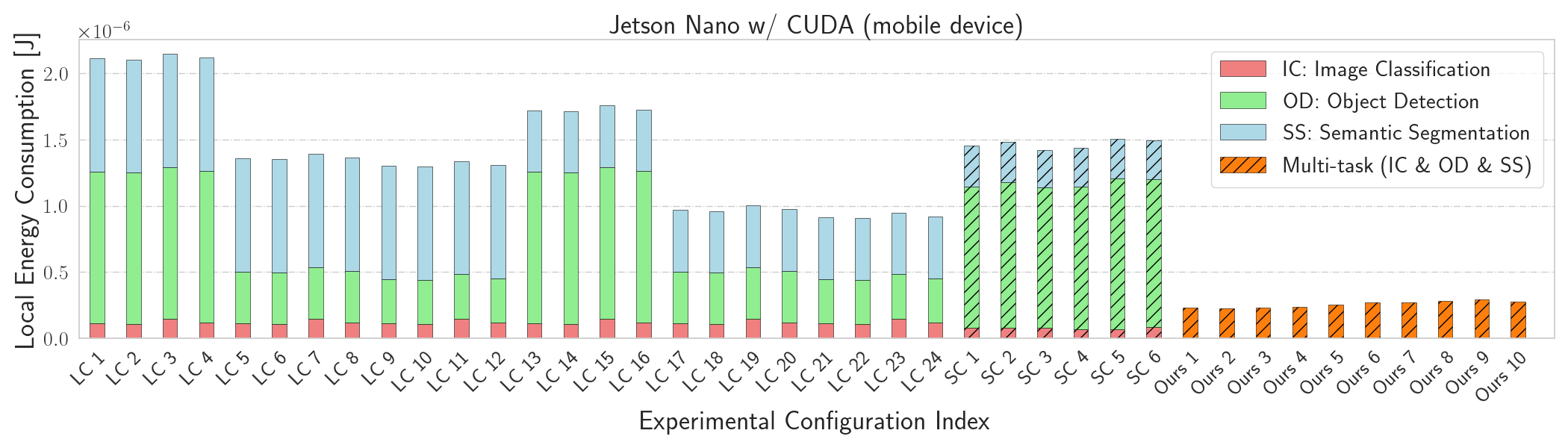}
    \vspace{-1em}
    \caption{Energy consumption of Jetson Nano (mobile device). Top/bottom: local computing without/with CUDA.}
    \label{fig:jetson_nano_w_cuda-energy_consumption}
    \vspace{1.5em}
    \includegraphics[width=0.975\linewidth]{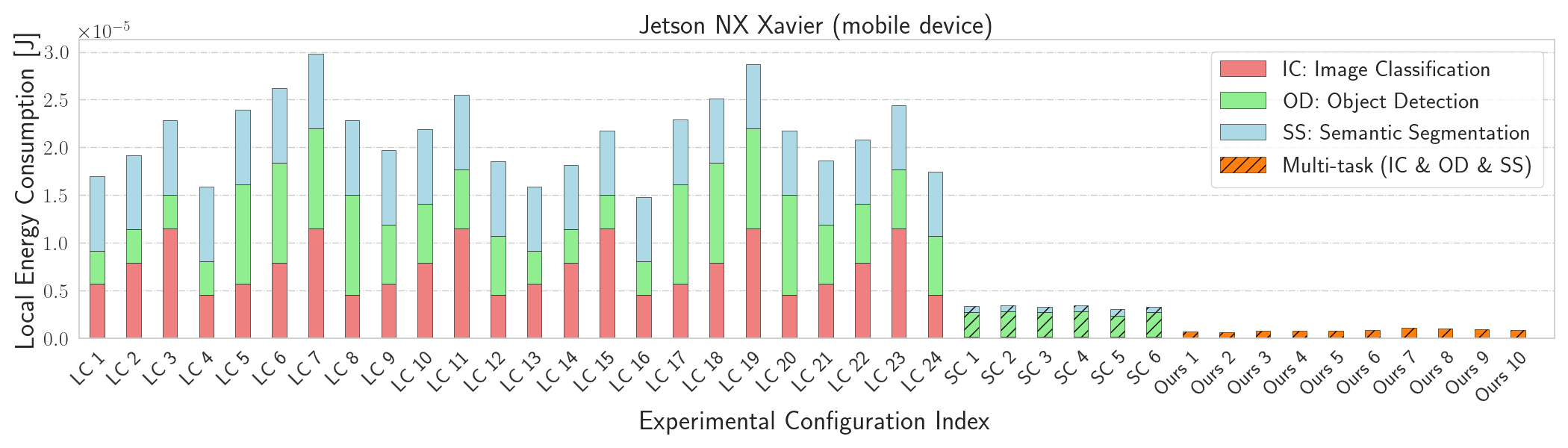}
    \label{fig:jetson_nx-energy_consumption}
    \vspace{1em}
    \includegraphics[width=0.975\linewidth]{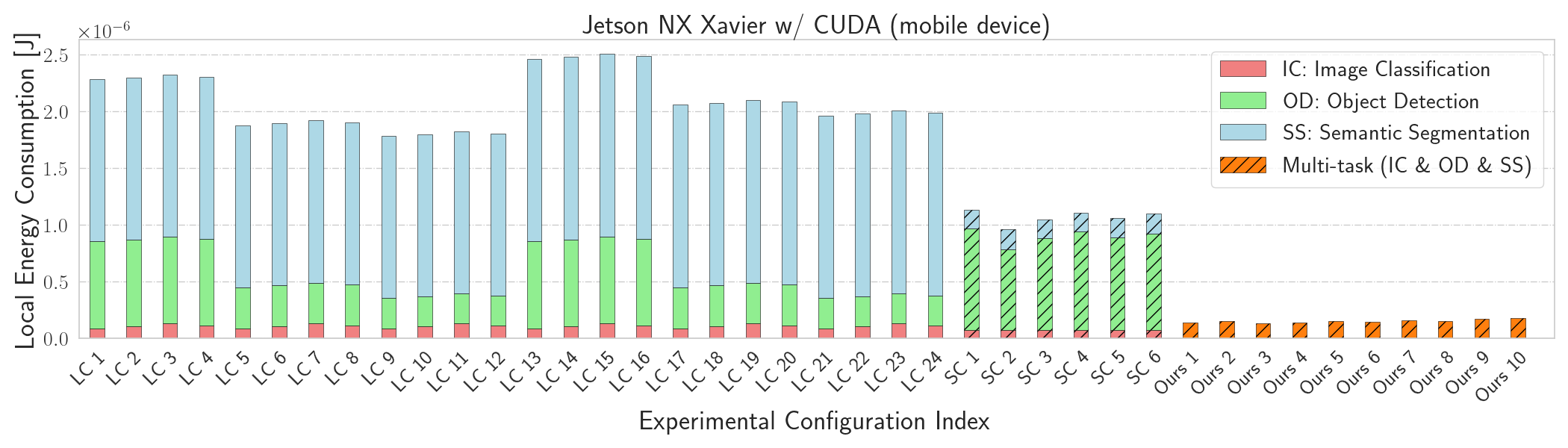}
    \vspace{-1em}
    \caption{Energy consumption of Jetson NX Xavier (mobile device). Top/bottom: local computing without/with CUDA.}
    \label{fig:jetson_nx_w_cuda-energy_consumption}
\end{figure*}

As illustrated in Figs.~\ref{fig:jetson_nano_w_cuda-energy_consumption} and~\ref{fig:jetson_nx_w_cuda-energy_consumption}, the overall trend is that the split computing approaches can significantly save energy consumption of mobile device in multi-task scenarios with respect to local computing even when the split computing approaches achieve higher end-to-end latency than the local computing approaches do (Figs.~\ref{fig:jetson_nano_w_cuda-laptop_w_cuda-100.0kbps} and~\ref{fig:jetson_nx_w_cuda-laptop_w_cuda-100.0kbps}).
Importantly, our \ourmodel models further saved energy consumption of the mobile devices.
Specifically, we reduced the energy consumption of the stronger baselines (split computing) by $65.0 - 88.2$\%, which highlighted the energy efficiency of our multi-task-head supervised compression models.

\section{Conclusions}
\label{sec:conclusion}

Split computing emerged in research communities as a promising solution for deep learning models in resource-constrained edge computing systems.
Existing studies on split computing are focused on task-specific models and assessments.
In this study, we proposed \ourmodel, the first end-to-end multi-task supervised compression model designed to share a unified image preprocessing pipeline and network architectures across multiple tasks for efficiently serving the multiple tasks in a single inference.

Using the SC2 benchmark and real devices, we comprehensively assessed its performances for three challenging computer vision tasks, end-to-end latency, and energy consumption of mobile devices in multi-task scenarios.
Learning compressed representations at early layers in a supervised manner for split computing, the \ourmodel models either outperformed or rivaled popular mobile-friendly models for ILSVRC 2012, COCO 2017, and PASCAL VOC 2012 in terms of predictive performance.
Overall, our models also outperformed LC and SC baselines in terms of end-to-end latency and energy consumption of mobile devices.
If energy consumption is not a concern, the LC baselines may be reasonable options as they achieved short latency in some cases at the cost of up to 8x energy consumption.

\section*{Acknowledgements}
This work was partially supported by the National Science Foundation under grants CNS 2134567 and CCF 2140154.

{\small
\bibliographystyle{ieee_fullname}
\bibliography{references}

\begin{thebibliography}{10}\itemsep=-1pt

\bibitem{agiollo2024enea}
Andrea Agiollo, Paolo Bellavista, Matteo Mendula, and Andrea Omicini.
\newblock {EneA-FL: Energy-aware orchestration for serverless federated learning}.
\newblock {\em Future Generation Computer Systems}, 154:219--234, 2024.

\bibitem{balle2017end}
Johannes Ball{\'e}, Valero Laparra, and Eero~P Simoncelli.
\newblock {End-to-end Optimized Image Compression}.
\newblock {\em International Conference on Learning Representations}, 2017.

\bibitem{balle2018variational}
Johannes Ball{\'e}, David Minnen, Saurabh Singh, Sung~Jin Hwang, and Nick Johnston.
\newblock {Variational image compression with a scale hyperprior}.
\newblock In {\em International Conference on Learning Representations}, 2018.

\bibitem{chen2017rethinking}
Liang-Chieh Chen, George Papandreou, Florian Schroff, and Hartwig Adam.
\newblock {Rethinking Atrous Convolution for Semantic Image Segmentation}.
\newblock {\em arXiv preprint arXiv:1706.05587}, 2017.

\bibitem{dao2014performance}
Thanh~Tuan Dao, Jungwon Kim, Sangmin Seo, Bernhard Egger, and Jaejin Lee.
\newblock {A Performance Model for GPUs with Caches}.
\newblock {\em IEEE Transactions on Parallel and Distributed Systems}, 26(7):1800--1813, 2014.

\bibitem{di2019design}
Marco Di~Vaio, Paolo Falcone, Robert Hult, Alberto Petrillo, Alessandro Salvi, and Stefania Santini.
\newblock {Design and Experimental Validation of a Distributed Interaction Protocol for Connected Autonomous Vehicles at a Road Intersection}.
\newblock {\em IEEE Transactions on Vehicular Technology}, 68(10):9451--9465, 2019.

\bibitem{eshratifar2019bottlenet}
Amir~Erfan Eshratifar, Amirhossein Esmaili, and Massoud Pedram.
\newblock {BottleNet: A Deep Learning Architecture for Intelligent Mobile Cloud Computing Services}.
\newblock In {\em 2019 IEEE/ACM Int. Symposium on Low Power Electronics and Design (ISLPED)}, pages 1--6, 2019.

\bibitem{everingham2012pascal}
Mark Everingham, Luc Van~Gool, CKI Williams, John Winn, and Andrew Zisserman.
\newblock {The PASCAL Visual Object Classes Challenge 2012 (VOC2012)}.
\newblock 2012.

\bibitem{he2019rethinking}
Kaiming He, Ross Girshick, and Piotr Doll{\'a}r.
\newblock {Rethinking ImageNet Pre-training}.
\newblock In {\em Proceedings of the IEEE/CVF International Conference on Computer Vision}, pages 4918--4927, 2019.

\bibitem{he2016deep}
Kaiming He, Xiangyu Zhang, Shaoqing Ren, and Jian Sun.
\newblock {Deep Residual Learning for Image Recognition}.
\newblock In {\em Proceedings of the IEEE Conference on Computer Vision and Pattern Recognition}, pages 770--778, 2016.

\bibitem{hinton2014distilling}
Geoffrey Hinton, Oriol Vinyals, and Jeff Dean.
\newblock {Distilling the Knowledge in a Neural Network}.
\newblock In {\em Deep Learning and Representation Learning Workshop: NIPS 2014}, 2014.

\bibitem{howard2019searching}
Andrew Howard, Mark Sandler, Grace Chu, Liang-Chieh Chen, Bo Chen, Mingxing Tan, Weijun Wang, Yukun Zhu, Ruoming Pang, Vijay Vasudevan, et~al.
\newblock {Searching for MobileNetV3}.
\newblock In {\em Proceedings of the IEEE/CVF International Conference on Computer Vision}, pages 1314--1324, 2019.

\bibitem{kingma2015adam}
Diederik~P. Kingma and Jimmy Ba.
\newblock {Adam: A Method for Stochastic Optimization}.
\newblock In {\em Third International Conference on Learning Representations}, 2015.

\bibitem{kingma2013auto}
Diederik~P Kingma and Max Welling.
\newblock {Auto-Encoding Variational Bayes}.
\newblock In {\em International Conference on Learning Representations}, 2014.

\bibitem{lin2017feature}
Tsung-Yi Lin, Piotr Doll{\'a}r, Ross Girshick, Kaiming He, Bharath Hariharan, and Serge Belongie.
\newblock {Feature Pyramid Networks for Object Detection}.
\newblock In {\em Proceedings of the IEEE Conference on Computer Vision and Pattern Recognition}, pages 2117--2125, 2017.

\bibitem{lin2014microsoft}
Tsung-Yi Lin, Michael Maire, Serge Belongie, James Hays, Pietro Perona, Deva Ramanan, Piotr Doll{\'a}r, and C~Lawrence Zitnick.
\newblock {Microsoft COCO: Common Objects in Context}.
\newblock In {\em European conference on computer vision}, pages 740--755. Springer, 2014.

\bibitem{liu2016ssd}
Wei Liu, Dragomir Anguelov, Dumitru Erhan, Christian Szegedy, Scott Reed, Cheng-Yang Fu, and Alexander~C Berg.
\newblock {SSD: Single shot multibox detector}.
\newblock In {\em European conference on computer vision}, pages 21--37, 2016.

\bibitem{liu2015deep}
Ziwei Liu, Ping Luo, Xiaogang Wang, and Xiaoou Tang.
\newblock {Deep Learning Face Attributes in the Wild}.
\newblock In {\em Proceedings of the IEEE international conference on computer vision}, pages 3730--3738, 2015.

\bibitem{matsubara2019distilled}
Yoshitomo Matsubara, Sabur Baidya, Davide Callegaro, Marco Levorato, and Sameer Singh.
\newblock {Distilled Split Deep Neural Networks for Edge-Assisted Real-Time Systems}.
\newblock In {\em Proceedings of the 2019 Workshop on Hot Topics in Video Analytics and Intelligent Edges}, pages 21--26, 2019.

\bibitem{matsubara2020head}
Yoshitomo Matsubara, Davide Callegaro, Sabur Baidya, Marco Levorato, and Sameer Singh.
\newblock {Head Network Distillation: Splitting Distilled Deep Neural Networks for Resource-Constrained Edge Computing Systems}.
\newblock {\em IEEE Access}, 8:212177--212193, 2020.

\bibitem{matsubara2020neural}
Yoshitomo Matsubara and Marco Levorato.
\newblock {Neural Compression and Filtering for Edge-assisted Real-time Object Detection in Challenged Networks}.
\newblock In {\em 2020 25th International Conference on Pattern Recognition (ICPR)}, pages 2272--2279, 2021.

\bibitem{matsubara2022split}
Yoshitomo Matsubara, Marco Levorato, and Francesco Restuccia.
\newblock {Split Computing and Early Exiting for Deep Learning Applications: Survey and Research Challenges}.
\newblock {\em ACM Computing Surveys}, 55(5):1--30, 2022.

\bibitem{matsubara2022supervised}
Yoshitomo Matsubara, Ruihan Yang, Marco Levorato, and Stephan Mandt.
\newblock {Supervised Compression for Resource-Constrained Edge Computing Systems}.
\newblock In {\em Proceedings of the IEEE/CVF Winter Conference on Applications of Computer Vision}, pages 2685--2695, 2022.

\bibitem{matsubara2023sc2}
Yoshitomo Matsubara, Ruihan Yang, Marco Levorato, and Stephan Mandt.
\newblock {SC2 Benchmark: Supervised Compression for Split Computing}.
\newblock {\em Transactions on Machine Learning Research}, 2023.

\bibitem{mcmahan2017communication}
Brendan McMahan, Eider Moore, Daniel Ramage, Seth Hampson, and Blaise~Aguera y Arcas.
\newblock {Communication-Efficient Learning of Deep Networks from Decentralized Data}.
\newblock In {\em Artificial intelligence and statistics}, pages 1273--1282. PMLR, 2017.

\bibitem{mendula2024furcifer}
Matteo Mendula, Paolo Bellavista, Marco Levorato, and Sharon~Ladron de Guevara~Contreras.
\newblock {Furcifer: a Context Adaptive Middleware for Real-world Object Detection Exploiting Local, Edge, and Split Computing in the Cloud Continuum}.
\newblock In {\em 2024 IEEE International Conference on Pervasive Computing and Communications (PerCom)}, pages 47--56. IEEE, 2024.

\bibitem{nimi2022chimera}
Sumaiya~Tabassum Nimi, Md~Adnan Arefeen, Md~Yusuf~Sarwar Uddin, Biplob Debnath, and Srimat Chakradhar.
\newblock {Chimera: Context-Aware Splittable Deep Multitasking Models for Edge Intelligence}.
\newblock In {\em 2022 IEEE International Conference on Smart Computing (SMARTCOMP)}, pages 70--77. IEEE, 2022.

\bibitem{paszke2019pytorch}
Adam Paszke, Sam Gross, Francisco Massa, Adam Lerer, James Bradbury, Gregory Chanan, Trevor Killeen, Zeming Lin, Natalia Gimelshein, Luca Antiga, et~al.
\newblock {PyTorch: An imperative style, high-performance deep learning library}.
\newblock In {\em Advances in Neural Information Processing Systems}, pages 8024--8035, 2019.

\bibitem{prometheus}
Bjorn Rabenstein and Julius Volz.
\newblock Prometheus: A next-generation monitoring system (talk).
\newblock Dublin, May 2015. {USENIX} Association.

\bibitem{ren2015faster}
Shaoqing Ren, Kaiming He, Ross Girshick, and Jian Sun.
\newblock {Faster R-CNN: Towards Real-Time Object Detection with Region Proposal Networks}.
\newblock In {\em Advances in Neural Information Processing Systems}, pages 91--99, 2015.

\bibitem{russakovsky2015imagenet}
Olga Russakovsky, Jia Deng, Hao Su, Jonathan Krause, Sanjeev Satheesh, Sean Ma, Zhiheng Huang, Andrej Karpathy, Aditya Khosla, Michael Bernstein, Alexander~C. Berg, and Li Fei-Fei.
\newblock {ImageNet Large Scale Visual Recognition Challenge}.
\newblock {\em International Journal of Computer Vision}, 115(3):211--252, 2015.

\bibitem{samie2016iot}
Farzad Samie, Lars Bauer, and J{\"o}rg Henkel.
\newblock {IoT Technologies for Embedded Computing: A Survey}.
\newblock In {\em 2016 International Conference on Hardware/Software Codesign and System Synthesis (CODES+ ISSS)}, pages 1--10. IEEE, 2016.

\bibitem{sandler2018mobilenetv2}
Mark Sandler, Andrew Howard, Menglong Zhu, Andrey Zhmoginov, and Liang-Chieh Chen.
\newblock {MobileNetV2: Inverted Residuals and Linear Bottlenecks}.
\newblock In {\em Proceedings of the IEEE Conference on Computer Vision and Pattern Recognition}, pages 4510--4520, 2018.

\bibitem{silberman2012indoor}
Nathan Silberman, Derek Hoiem, Pushmeet Kohli, and Rob Fergus.
\newblock {Indoor Segmentation and Support Inference from RGBD Images}.
\newblock In {\em Computer Vision -- ECCV 2012}, pages 746--760. Springer, 2012.

\bibitem{singh2020end}
Saurabh Singh, Sami Abu-El-Haija, Nick Johnston, Johannes Ball{\'e}, Abhinav Shrivastava, and George Toderici.
\newblock {End-to-end Learning of Compressible Features}.
\newblock In {\em 2020 IEEE International Conference on Image Processing (ICIP)}, pages 3349--3353. IEEE, 2020.

\bibitem{tallarida1987area}
Ronald~J Tallarida, Rodney~B Murray, Ronald~J Tallarida, and Rodney~B Murray.
\newblock {Area under a Curve: Trapezoidal and Simpson's Rules}.
\newblock {\em Manual of Pharmacologic Calculations: with Computer Programs}, pages 77--81, 1987.

\bibitem{tan2019mnasnet}
Mingxing Tan, Bo Chen, Ruoming Pang, Vijay Vasudevan, Mark Sandler, Andrew Howard, and Quoc~V Le.
\newblock {MnasNet: Platform-Aware Neural Architecture Search for Mobile}.
\newblock In {\em Proceedings of the IEEE Conf. on Computer Vision and Pattern Recognition}, pages 2820--2828, 2019.

\bibitem{vepakomma2018split}
Praneeth Vepakomma, Otkrist Gupta, Tristan Swedish, and Ramesh Raskar.
\newblock {Split learning for health: Distributed deep learning without sharing raw patient data}.
\newblock {\em arXiv preprint arXiv:1812.00564}, 2018.

\bibitem{zhang2022resnest}
Hang Zhang, Chongruo Wu, Zhongyue Zhang, Yi Zhu, Haibin Lin, Zhi Zhang, Yue Sun, Tong He, Jonas Mueller, R Manmatha, et~al.
\newblock {ResNeSt: Split-Attention Networks}.
\newblock In {\em Proceedings of the IEEE/CVF conference on computer vision and pattern recognition}, pages 2736--2746, 2022.

\end{thebibliography}
}

\renewcommand{\thetable}{S\Roman{table}}
\renewcommand{\thefigure}{S\arabic{figure}}
\renewcommand{\theequation}{S\arabic{equation}}

\newpage
\section*{Supplementary Material}
\appendix
\section{Hyperparameters for \ourmodel model training}
\label{sec:model_training}
\subsection*{Step 1: Pre-training encoder-decoder}
We pre-train the encoder-decoder modules of our \ourmodel models with the Adam optimizer~\cite{kingma2015adam}, minimizing the loss function defined in Eq. (\ref{eq:1st_loss}) in the main paper.
We use outputs of the first, second, third, and fourth residual / ResNeSt blocks from the pre-trained ResNet-50 as $\h_i^{t}(\x)$ and extract those of the corresponding residual blocks in our ResNet-50-based \ourmodel models as $\h_i^{s}(\x)$.
$\beta$ is a hyperparameter to control the rate-distortion tradeoff.
Given a \ourmodel architecture, we train five individual models with $\beta = 0.32, 0.64, 1.28, 2.56,$ and $5.12$.
The same procedure is applied when teacher and student models are based on ResNeSt-269e~\cite{zhang2022resnest}.
We use the trainind set of the ILSVRC 2012 dataset~\cite{russakovsky2015imagenet} for 10 epochs, and training batch size is 32.
The initial learning rate is $0.001$ and exponentially decayed by a factor of $0.1$ after the first 5 and 8 epochs.

\subsection*{Step 2: Fine-tuning decoder and subsequent modules}
Following Step 1, we freeze parameters of the encoder and entropy bottleneck.
We then fine-tune the remaining modules including a classification head as illustrated in Fig.~\ref{fig:es_vs_ladon} (bottom).
Specifically, we fine-tune the modules for 10 epochs using the pretrained ResNet-50 (ResNeSt-269e) as a teacher model for a standard knowledge distillation (KD) loss function~\cite{hinton2014distilling}
\begin{equation}
    \mathcal{L} = \alpha \cdot \text{CE}(\mathbf{\hat{y}}, \mathbf{y}) + (1 - \alpha) \cdot \tau^2 \cdot \text{KL} \left(\mathbf{o}^\text{S}, \mathbf{o}^\text{T}\right),
    \label{eq:proposed_2nd}
\end{equation}
\noindent where \text{CE} and \text{KL} are cross-entropy and Kullbuck-Leibler divergence, respectively.
$\mathbf{\hat{y}}$ and $\mathbf{y}$ are true and predicted class labels.
$\alpha \in [0, 1]$ and $\tau$ are hyperparameters.
We use $\alpha = 0.5$ and $\tau = 1$ in this study.
$\mathbf{o}^\text{T}$ and $\mathbf{o}^\text{S}$ indicate\emph{softened} output distributions produced by teacher and student models, respectively.
$\mathbf{o}^\text{T} = [o_1^\text{T}, o_2^\text{T}, \ldots, o_{|\mathcal{C}|}^\text{T}]$ where $\mathcal{C}$ is a set of object categories in the target task, which is an image classification for the ILSVRC 2012 dataset~\cite{russakovsky2015imagenet}.
$o_i^\text{T}$ is the teacher model's softened output value (scalar) for the $i$-th object category:
\begin{equation}
    o_{i}^\text{T} = \frac{\exp \left( \frac{v_i^\text{T}}{\tau} \right)}{\sum_{k \in \mathcal{C}} \exp\left( \frac{v_k^\text{T}}{\tau} \right)},
\end{equation}
\noindent where $v_i^\text{T}$ is the teacher model's logit value for the $i$-th object category.
The same rule is applied to the student model.

Here, we use a stochastic gradient descent (SGD) optimizer with the initial learning rate of $0.001$, momentum of $0.9$, and weight decay of $0.0005$.
The learning rate is exponentially decayed by a factor of $0.1$ after the first 5 epochs.
At the end of this step, all the modules in our \ourmodel model required for the image classification task are ready to serve.

\subsection*{Step 3: Fine-tuning other task-specific modules}
Following Step 2, we freeze all the learnt parameters and then introduce other task-specific modules.

\paragraph{Object Detection}
We introduce to the fine-tuned \ourmodel model, an object detection head that consists of Faster R-CNN~\cite{ren2015faster} and Feature Pyramid Network (FPN)~\cite{lin2017feature} modules.
We train the object detection head on COCO 2017~\cite{lin2014microsoft} for 26 epochs (ResNet-based \ourmodel) and 28 epochs (ResNeSt-based \ourmodel), minimizing a linear combination of bounding box regression, objectness, and object classification losses~\cite{ren2015faster}.
The SGD optimizer uses the initial learning rate of $0.02$, momentum of $0.9$, weight decay of $0.0001$, and batch size of $8$.
Its learning rate is exponentially decayed by a factor of $0.1$ after the first 16 and 22 epochs.
For ResNeSt-based \ourmodel, we use the weight decay of $0.0005$ and exponentially decay the learning rate by a factor of $0.1$ after the first 10, 18 and 24 epochs.

\begin{figure*}[h]
    \centering
    \includegraphics[width=0.975\linewidth]{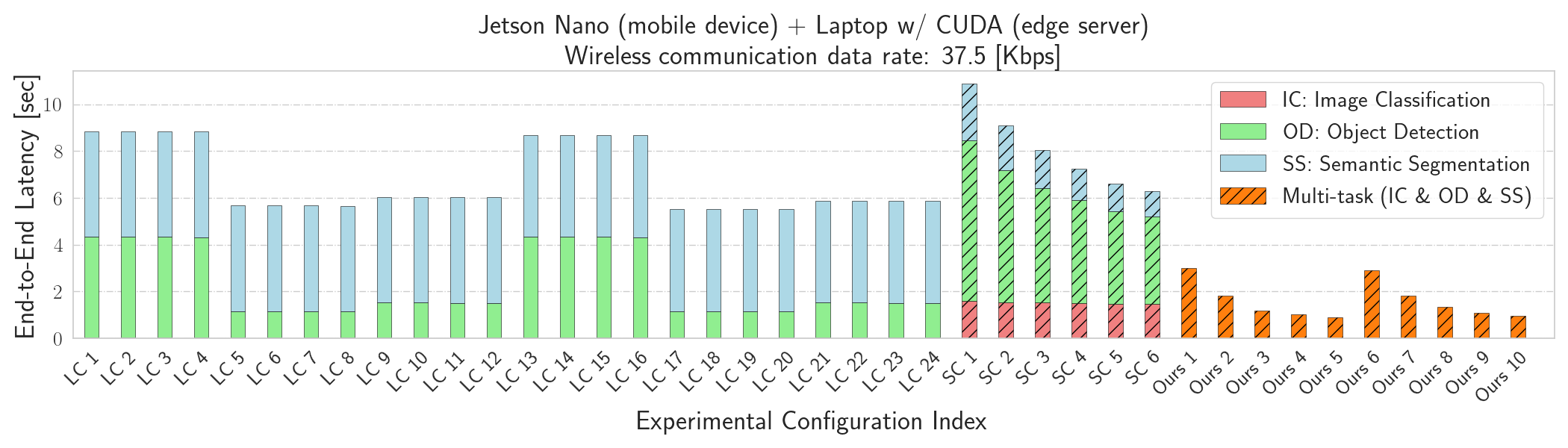}
    \label{fig:jetson_nano-laptop_w_cuda-37.5kbps}
    \vspace{1em}
    \includegraphics[width=0.975\linewidth]{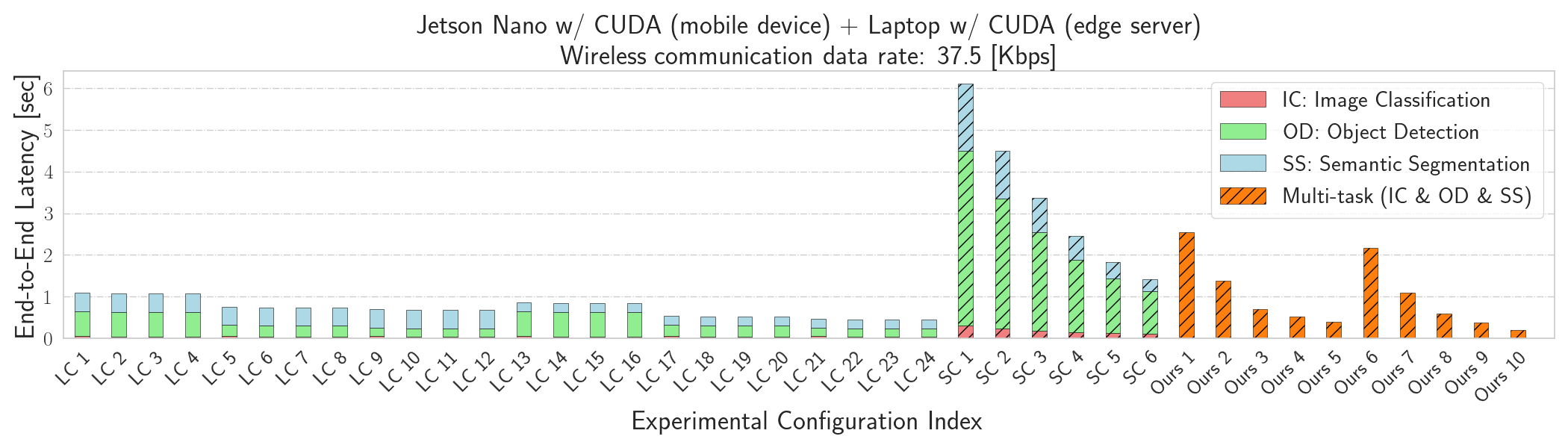}
    \vspace{-1em}
    \caption{End-to-end latency for Jetson Nano (mobile device), laptop with CUDA (edge server), and wireless communication data rate of 37.5 Kbps. Top/bottom: local computing without/with CUDA.}
    \label{fig:jetson_nano_w_cuda-laptop_w_cuda-37.5kbps}
    \vspace{1em}
    \includegraphics[width=0.975\linewidth]{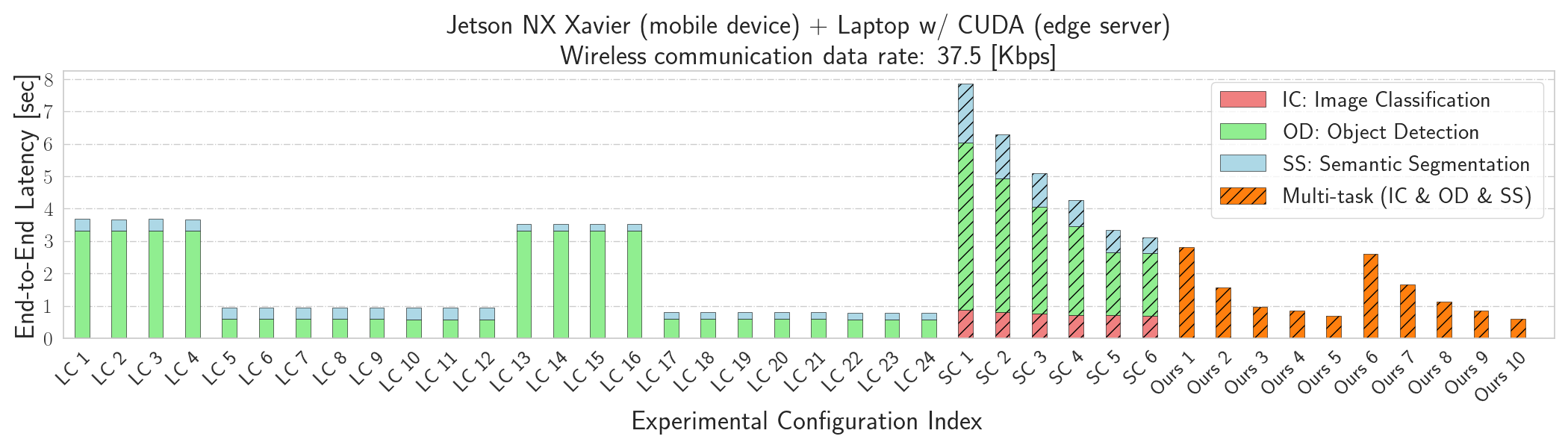}
    \label{fig:jetson_nx-laptop_w_cuda-37.5kbps}
    \vspace{1em}
    \includegraphics[width=0.975\linewidth]{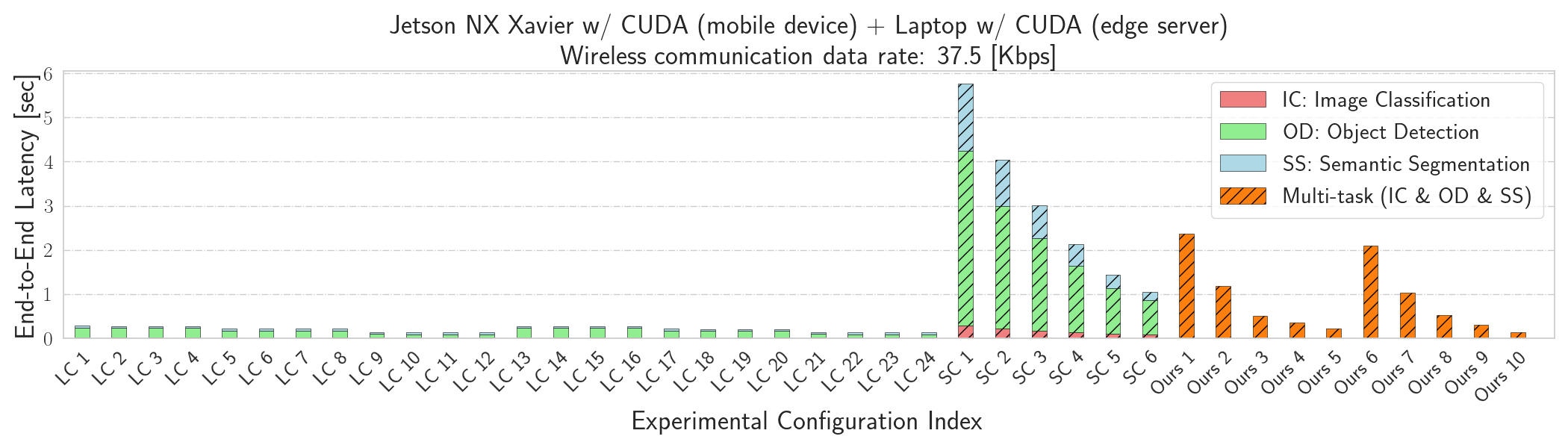}
    \vspace{-1em}
    \caption{End-to-end latency for Jetson NX Xavier (mobile device), laptop with CUDA (edge server), and wireless communication data rate of 37.5 Kbps. Top/bottom: local computing without/with CUDA.}
    \label{fig:jetson_nx_w_cuda-laptop_w_cuda-37.5kbps}
\end{figure*}

\begin{figure*}[t]
    \centering
    \includegraphics[width=0.975\linewidth]{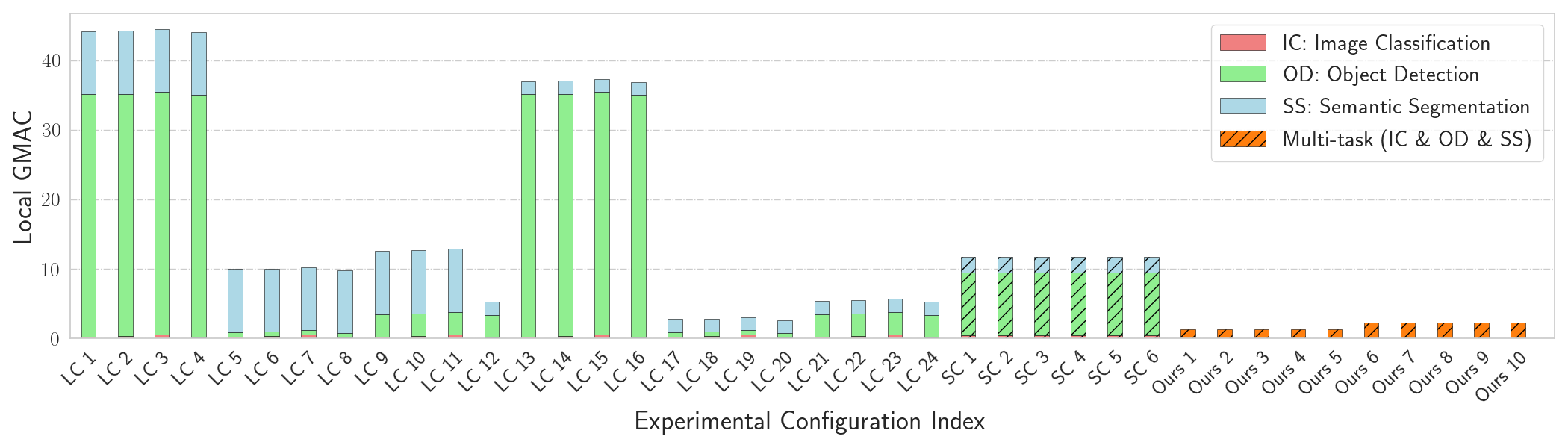}
    \vspace{-1em}
    \caption{Local Giga Multiply-Accumulate Operation (GMAC).}
    \label{fig:local_gmac}
\end{figure*}

\paragraph{Semantic Segmentation}
For a semantic segmentation head, we employ DeepLabv3~\cite{chen2017rethinking}.
With the SGD optimizer, we train the semantic segmentation head by minimizing the standard cross-entropy for 90 epochs, using momentum of $0.9$, weight decay of $0.0001$, and batch size of 16.
For the first 30 epochs, we use COCO 2017 training dataset~\cite{lin2014microsoft} with the initial learning rates are $0.02$ and $0.01$ for the semantic segmentation head and its auxiliary classifier, respectively.
We follow~\cite{chen2017rethinking} and reduce the learning rates at every iteration
\begin{equation}
    \eta = \eta_0 \times \left(1 - \frac{t}{N_\text{iter}}\right)^{0.9},
    \label{eq:seg_lr}
\end{equation}
\noindent where $\eta_0$ is the initial learning rate.
$t$ and $N_\text{iter}$ are the current iteration count and the total number of iterations, respectively.

For the last 60 epochs, we use the PASCAL VOC 2012 training dataset~\cite{everingham2012pascal} and fine-tune the semantic segmentation head.
Other hyperparameters are the same as those for the first 30 epochs.

\section{End-to-end Latency with LoRa}
\label{subsec:e2e_latency_lora}

In the main paper (Figs.~\ref{fig:jetson_nano_w_cuda-laptop_w_cuda-100.0kbps}) and~\ref{fig:jetson_nx_w_cuda-laptop_w_cuda-100.0kbps}), we show end-to-end latency evaluations using a challenged wireless network, where we assume the data rate is only $100$ Kbps.
Here, we consider LoRa~\cite{samie2016iot}, a further challenged network condition whose maximum data rate is $37.5$ Kbps, and perform another end-to-end latency evaluation using the same experimental configurations (see Table~\ref{table:e2e_configs}).

Figures~\ref{fig:jetson_nano_w_cuda-laptop_w_cuda-37.5kbps} and~\ref{fig:jetson_nx_w_cuda-laptop_w_cuda-37.5kbps} show the end-to-end latency evaluation results for Jetson Nano and Jetson NX Xavier (top: CUDA OFF, bottom: CUDA ON) as mobile devices, respectively.
Note that the new experimental configuration does not affect the performance of the local computing (LC) baselines in Figs.~\ref{fig:jetson_nano_w_cuda-laptop_w_cuda-100.0kbps} and~\ref{fig:jetson_nx_w_cuda-laptop_w_cuda-100.0kbps} since the LC baselines do not offload computation to the edge server.
In other words, the configuration makes it more difficult for the SC baselines and our proposed method to outperform the LC baselines as the lower communication data rate will further delay communications between mobile devices and edge (cloud) servers.

The overall trends in Fig.~\ref{fig:jetson_nano_w_cuda-laptop_w_cuda-37.5kbps} are similar to those in Fig.~\ref{fig:jetson_nano_w_cuda-laptop_w_cuda-100.0kbps} in the main paper.
Our multi-task models (\ourmodel) saved up to $90.1$\% and $96.7$\% of the end-to-end latency with the LC and SC baselines, respectively.
Using Jetson NX Xavier in this scenario (Fig.~\ref{fig:jetson_nx_w_cuda-laptop_w_cuda-37.5kbps}) made the LC baselines even stronger.
While giving the LC baselines more advantage, the \ourmodel models reduced the end-to-end latency of the LC and SC baselines by up to $83.8$\% and $97.6$\% respectively.

\section{Local GMAC and Peak Local Memory Usage}
In this section, we briefly discuss computational loads on mobile device using local GMAC (Giga Multiply-Accumulate Operation) and peak local memory usage metrics.
As shown in Figure~\ref{fig:local_gmac}, our approach consistently achieved lower GMAC on mobile devices than baseline methods considered in this study, saving up to $97.0\%$ and $88.7\%$ of local GMAC for the LC and SC baselines, respectively.

Table~\ref{table:peak_local_memory} presents peak memory usage measured during our multi-task experiments.
Note that the reported memory consumption may include those of background jobs running on the mobile devices.
Overall, our approach improved peak memory usage on mobile devices over the baseline methods.

\begin{table}[h]
    \centering
    \small
    \bgroup
    \def\arraystretch{0.88}
    \begin{tabular}{l|rrrr}
        \toprule
        \multicolumn{1}{c|}{\multirow{2}{*}{\bf Config.}} & \multicolumn{2}{c}{\bf Jetson Nano} & \multicolumn{2}{c}{\bf Jetson NX Xavier} \\
        & \multicolumn{1}{c}{w/o CUDA} & \multicolumn{1}{c}{w/ CUDA} & \multicolumn{1}{c}{w/o CUDA} & \multicolumn{1}{c}{w/ CUDA} \\
        \hline\midrule
        LC 1 & 2.17 & 0.785 & 4.49 & 2.93 \\
        LC 2 & 2.17 & 0.651 & 4.49 & 2.93 \\
        LC 3 & 2.17 & 0.632 & 4.49 & 2.93 \\
        LC 4 & 2.17 & 0.632 & 4.49 & 3.10 \\
        LC 5 & 2.17 & 0.785 & 4.49 & 2.93 \\
        LC 6 & 2.17 & 0.651 & 4.49 & 2.93 \\
        LC 7 & 2.17 & 0.632 & 4.49 & 2.93 \\
        LC 8 & 2.17 & 0.632 & 4.49 & 3.10 \\
        LC 9 & 2.17 & 0.785 & 4.49 & 2.93 \\
        LC 10 & 2.17 & 0.651 & 4.49 & 2.93 \\
        LC 11 & 2.17 & 0.632 & 4.49 & 2.93 \\
        LC 12 & 2.17 & 0.632 & 4.49 & 3.10 \\
        LC 13 & 2.16 & 1.11 & 4.36 & 3.65 \\
        LC 14 & 2.16 & 1.11 & 4.36 & 3.65 \\
        LC 15 & 2.16 & 1.11 & 4.36 & 3.65 \\
        LC 16 & 2.16 & 1.11 & 4.36 & 3.65 \\
        LC 17 & 2.16 & 1.11 & 4.36 & 3.65 \\
        LC 18 & 2.16 & 1.11 & 4.36 & 3.65 \\
        LC 19 & 2.16 & 1.11 & 4.36 & 3.65 \\
        LC 20 & 2.16 & 1.11 & 4.36 & 3.65 \\
        LC 21 & 2.16 & 1.11 & 4.36 & 3.65 \\
        LC 22 & 2.16 & 1.11 & 4.36 & 3.65 \\
        LC 23 & 2.16 & 1.11 & 4.36 & 3.65 \\
        LC 24 & 2.16 & 1.11 & 4.36 & 3.65 \\
        \midrule
        SC 1 & 1.16 & 0.678 & 3.21 & 3.14 \\
        SC 2 & 0.937 & 0.618 & 3.08 & 3.63 \\
        SC 3 & 0.798 & 0.712 & 3.16 & 3.54 \\
        SC 4 & 0.760 & 0.699 & 2.91 & 3.60 \\
        SC 5 & 0.774 & 0.738 & 2.85 & 3.31 \\
        SC 6 & 0.790 & 0.689 & 2.72 & 3.64 \\
        \midrule
        Ours 1 & 0.409 & 0.560 & 0.868 & 2.12 \\
        Ours 2 & 0.502 & 0.535 & 0.599 & 1.94 \\
        Ours 3 & 0.459 & 0.524 & 0.737 & 2.30 \\
        Ours 4 & 0.436 & 0.563 & 0.396 & 2.17 \\
        Ours 5 & 0.439 & 0.539 & 0.450 & 2.03 \\
        Ours 6 & 0.510 & 0.557 & 1.60 & 1.70 \\
        Ours 7 & 0.717 & 0.528 & 1.44 & 1.67 \\
        Ours 8 & 0.721 & 0.427 & 1.36 & 1.51 \\
        Ours 9 & 0.647 & 0.597 & 0.871 & 2.21 \\
        Ours 10 & 0.682 & 0.543 & 1.12 & 2.14 \\
        \bottomrule
    \end{tabular}
    \egroup
    \vspace{-0.5em}
    \caption{Peak local memory usage [GB] during multi-task experiments. Reported numbers may include memory consumption by background jobs on mobile devices.}
    \label{table:peak_local_memory}
\end{table}

\end{document}